\documentclass{midl} 

\usepackage{todonotes}
\usepackage{blindtext}
\usepackage{stackengine}
\usepackage{graphicx}
\usepackage{float}
\usepackage{subcaption}
\usepackage{mathtools}
\usepackage{booktabs}
\usepackage{enumitem}
\usepackage{mwe} 
\jmlrvolume{-- Under Review}
\jmlryear{2019}
\jmlrworkshop{Full Paper -- MIDL 2019 submission}
\newcommand{\mypm}{\mathbin{\smash{%
\raisebox{0.35ex}{%
            $\underset{\raisebox{0.5ex}{$\smash -$}}{\smash+}$%
            }%
        }%
    }%
}

\editors{Under Review for MIDL 2019}
\DeclareMathOperator{\E}{\mathbb{E}}
\title[Adversarial Pseudo Healthy Synthesis Needs Pathology Factorization]{Adversarial Pseudo Healthy Synthesis Needs Pathology Factorization}
\DeclarePairedDelimiter\norm{\lVert}{\rVert}

\midlauthor{\Name{Tian Xia\midlotherjointauthor\nametag{$^{1}$}} \Email{Tian.Xia@ed.ac.uk}\AND
\Name{Agisilaos Chartsias\midlotherjointauthor\nametag{$^{1}$}} \Email{agis.chartsias@ed.ac.uk}\AND
\Name{Sotirios A. Tsaftaris\midlotherjointauthor\nametag{$^{1, 2}$}} \Email{S.Tsaftaris@ed.ac.uk}\\
\addr $^{1}$ School of Engineering, University of Edinburgh, West Mains Rd, Edinburgh EH9 3FB, UK \\
$^{2}$ The Alan Turing Institute, London, UK
\\
}

\begin{document}

\maketitle
\begin{abstract}

Pseudo healthy synthesis, i.e. the creation of a subject-specific `healthy' image from a pathological one, could be helpful in tasks such as anomaly detection, understanding changes induced by pathology and disease or even as data augmentation. We treat this task as a factor decomposition problem:  we aim to separate what appears to be healthy and where disease is (as a map). The two factors are then recombined (by a network) to reconstruct the input disease image. We train our models in an adversarial way using either paired or unpaired settings, where we pair disease images and maps (as segmentation masks) when available. We quantitatively evaluate the quality of pseudo healthy images. We show in a series of experiments, performed in ISLES and BraTS datasets, that our method is better than conditional GAN and CycleGAN, highlighting challenges in using adversarial methods in the image translation task of pseudo healthy image generation.
\end{abstract}


\section{Introduction}



The aim of pseudo healthy synthesis is to synthesize a subject-specific `healthy' image from a pathological one. Generating such images can be valuable both in research and in clinical applications. For example, these images can be used as a means to perform pathology segmentation \cite{bowles2017brain, ye2013modality}, detection \cite{tsunoda2014pseudo}, to help with the visual understanding of disease classification networks \cite{baumgartner2017visual} and to aid experts with additional diagnostic information \cite{sun2018adversarial}.

A challenge with pseudo healthy synthesis is the lack of paired pathological and healthy images for training, i.e. we do not have images of the same patient moments before and after pathology has appeared. Thus, methods based on pure supervised learning are not fit for purpose. While longitudinal observations could perhaps partially alleviate this problem, the time difference between observations is an additional factor that may complicate learning. Thus, it is imperative to overcome this lack of paired data. One approach is to learn distributions that characterize the domains of healthy and pathological images, for example by learning a compact manifold of patch-based dictionaries \cite{ye2013modality, tsunoda2014pseudo}, or alternatively by learning mappings between the two domains with the use of adversarial training \cite{sun2018adversarial}. 

\begin{figure}[t]
\centering
\subcaptionbox{Example results \label{fig: example results1}} {\includegraphics[scale=0.2]{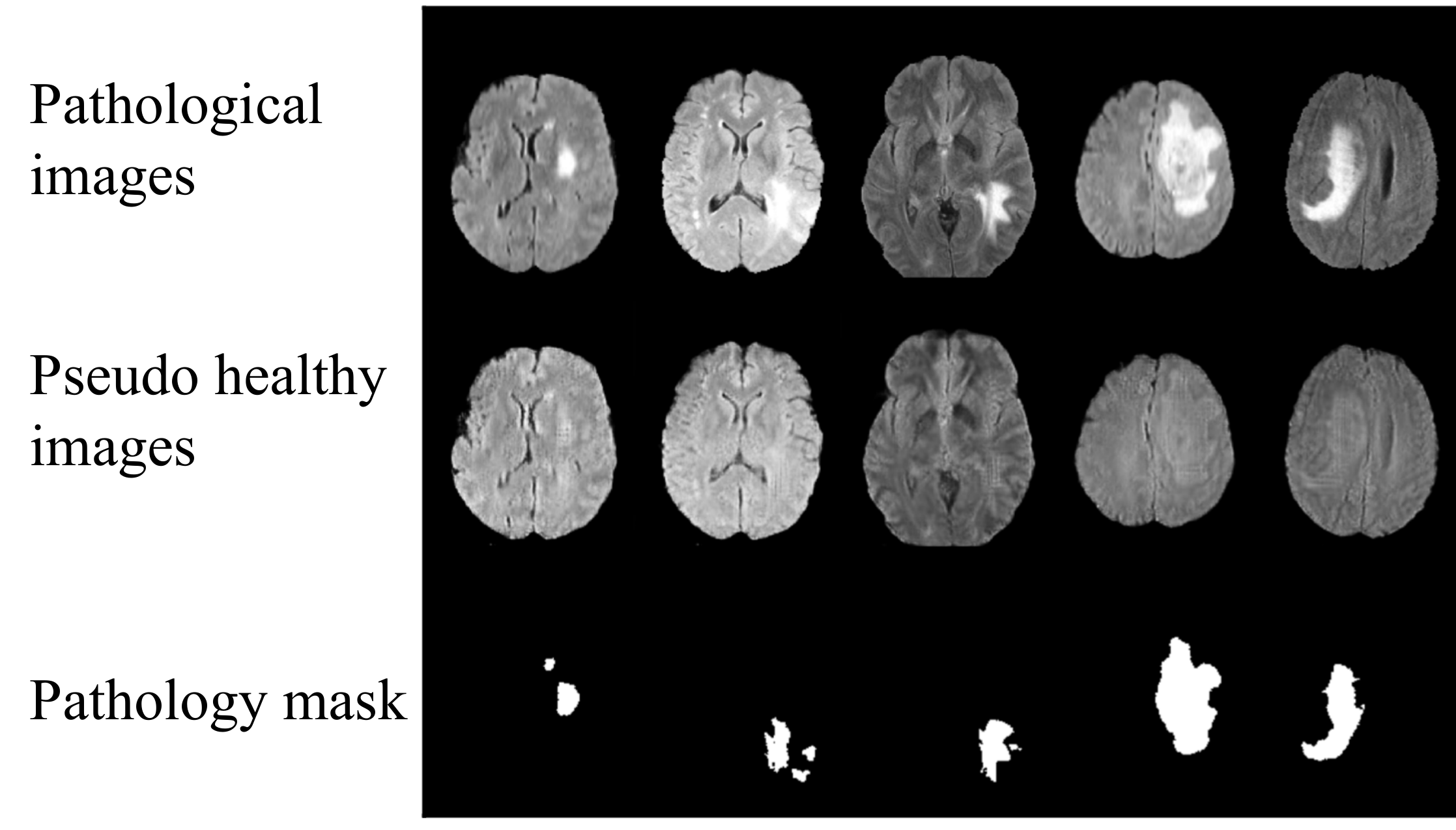}}%
\subcaptionbox{Simple illustration of our method \label{fig: example results2}} {\includegraphics[scale=0.4]{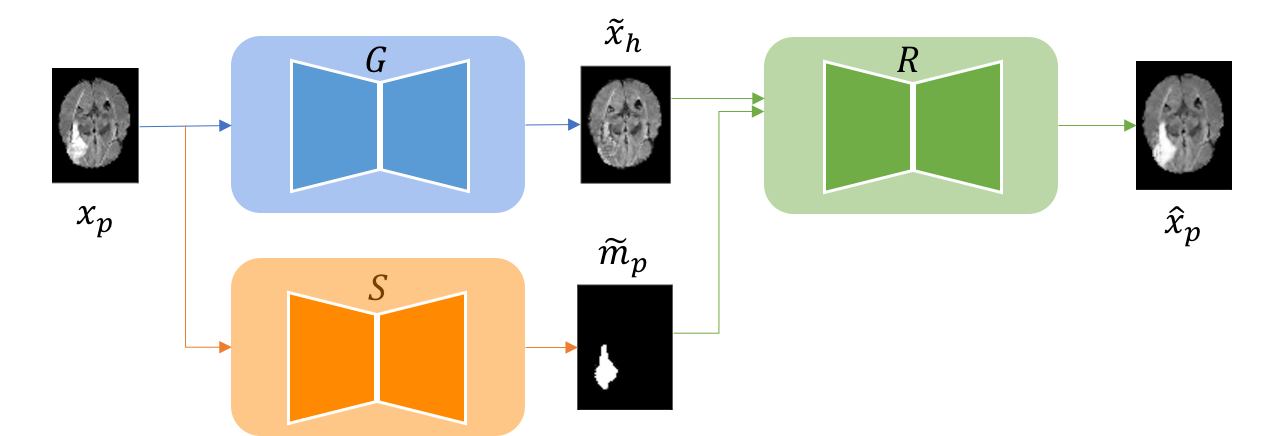}}%
\caption{Example results and simple illustration of our method. The three rows of (a) show input pathological images, corresponding pseudo healthy images, and pathology segmentation masks, respectively. Images are taken from the BraTS dataset. In (b) a pseudo healthy image $\tilde{x}_h$ and a pathology mask $\tilde{m}_p$ are generated from a pathological image $x_p$, and then finally a reconstructed image $\hat{x}_p$ is generated from $\tilde{x}_h$ and $\tilde{m}_p$. }
\label{fig: example results}
\end{figure}

We follow a similar approach here but 
focus on factorizing the pathology. Simple schematic and examples are shown in Figure \ref{fig: example results}. We aim to separate what appears to be healthy out of a disease image. We let neural networks decompose an input image into a healthy image (one factor), via a generator, and a binary map that aims to localize disease (the other factor) via a segmentor. These two factors are then composed together to reconstruct the input via another network. 
The pathological map is necessary as a factor to solve the one-to-many problem\footnote{There could be many disease images that could originate from the same healthy image, e.g. consider the simple setting of a lesion in many different locations on the same brain.} \cite{chu2017cyclegan}: the healthy image must by definition contain `less information' than the disease image.

We can train the segmentor in a supervised way using `\textit{paired}' pathological images and their corresponding masks. However, since annotations of pathology are not easy to acquire, we also propose an `\textit{unpaired}' training strategy. We take advantage of several losses including a cycle-consistency loss \cite{zhu2017unpaired}, but use a modified second cycle where we enforce healthy-to-healthy image translation to approach the identity. Finally, since most pseudo healthy methods focus on applications of the synthetic data, results are either evaluated qualitatively or by demonstrating improvements on downstream tasks. A direct quantitative evaluation of the quality of pseudo healthy images has been largely ignored. We propose two numerical evaluation metrics for characterizing the `\textit{healthiness}' (i.e. how close to being healthy) and `\textit{identity}' (i.e. how close to corresponding to the input identity) of synthetic results.

Our \textbf{contributions} in this work are three-fold:
\begin{enumerate}[topsep=0pt,itemsep=-1ex,partopsep=1ex,parsep=1ex]
  \item We propose a method that factorizes anatomical and pathological information.
  \item We consider two training settings:  (a) \textit{paired}: when we have paired images and ground-truth pathology masks;  (b) \textit{unpaired}: when such pairs are not available.
  \item We propose numerical evaluation metrics to explicitly evaluate the quality of pseudo healthy synthesized images, and compare our method with conditional GAN \cite{mirza2014conditional} and CycleGAN \cite{zhu2017unpaired} on ISLES and BraTS datasets.
\end{enumerate}

\section{Related work on pseudo healthy synthesis}
\noindent Medical image synthesis is an active research topic in medical image analysis \cite{special2018Frangi}
with an active community and dedicated workshops (e.g.\ the SASHIMI MICCAI series).  For brevity here we focus on methods related to pseudo healthy image synthesis with adversarial mechanisms. Image synthesis (translation) can be solved by a conditional GAN that learns a mapping between image domains (e.g.\ A to B).  However, preservation of `identity' is not guaranteed: there are no explicit costs to enforce that an image from domain A to be translated to the same image in domain B. 
%
CycleGAN uses a cycle-consistency loss to promote identity and has been profoundly adopted in medical image analysis \cite{huo2018, zhang2018translating, wolterink2017deep, chartsias2017adversarial, wang2018unsupervised}.

\citet{baumgartner2017visual} used Wasserstein GAN \cite{arjovsky2017wasserstein} to generate disease effect maps, and used these maps to synthesize pathological images. \citet{andermatt2018pathology} combined the idea of \citet{baumgartner2017visual} with CycleGAN to perform pseudo healthy synthesis for pathology detection. \citet{yang2016registration} used a Variational Auto-encoder to learn a mapping from pathological images to quasi-normal (pseudo healthy) images to improve atlas-to-image registration accuracy with large pathologies. 
\citet{schlegl2017unsupervised} and \citet{chen2018unsupervised} trained adversarial auto-encoder networks only on normal data, then used the trained model to synthesize normal data from abnormal data as a way of detecting the anomaly. 
\citet{sun2018adversarial} proposed a CycleGAN-based method to perform pseudo healthy synthesis treating  `pathological' and `healthy' as two domains. 

The majority of these works use pseudo healthy images to achieve improvements in downstream tasks. While the performance on such downstream tasks relies on pseudo healthy image quality, it is not explicitly evaluated. Herein, we pay particular attention to consistently evaluate how `healthy' the synthetic images look, and whether they correspond to the same `identity' of the input. 
All methods rely on some form of adversarial training to approximate a distribution. However, as we will detail below, when one of the domains has less information the one-to-many problem can appear and CycleGAN may collapse.
Our method treats pathology as a `residual' factor: it factorizes anatomical and pathological information using adversarial and cycle-consistent losses to bypass the one-to-many problem. 

\section{Methodology}
\subsection{Problem overview}
We denote a pathological image as \textit{x\textsubscript{p}} and a healthy image as \textit{x\textsubscript{h}}, drawn from $\mathcal{P}$ and $\mathcal{H}$ distributions, respectively, i.e $\textit{x\textsubscript{p}}\sim\mathcal{P}$ and $\textit{x\textsubscript{h}}\sim\mathcal{H}$. Our task is to generate a pseudo healthy image $\textit{$\mathit{\tilde{x}}$\textsubscript{h}}$ for a sample \textit{x\textsubscript{p}}, such that $\textit{$\mathit{\tilde{x}}$\textsubscript{h}}$ lies in the distribution of healthy images, i.e. $\textit{$\mathit{\tilde{x}}$\textsubscript{h}}\sim\mathcal{H}$. In the meantime, we also want the generated image $\textit{$\mathit{\tilde{x}}$\textsubscript{h}}$ to maintain the identity of the original image \textit{x\textsubscript{p}}, i.e. to come from the same subject as \textit{x\textsubscript{p}}. Therefore, pseudo healthy synthesis can be formulated as two major objectives: \textit{remove} the disease of pathological images, and \textit{maintain} the identity and realism as good as possible. 
\begin{figure}[t]
\centering
\includegraphics[scale=0.3]{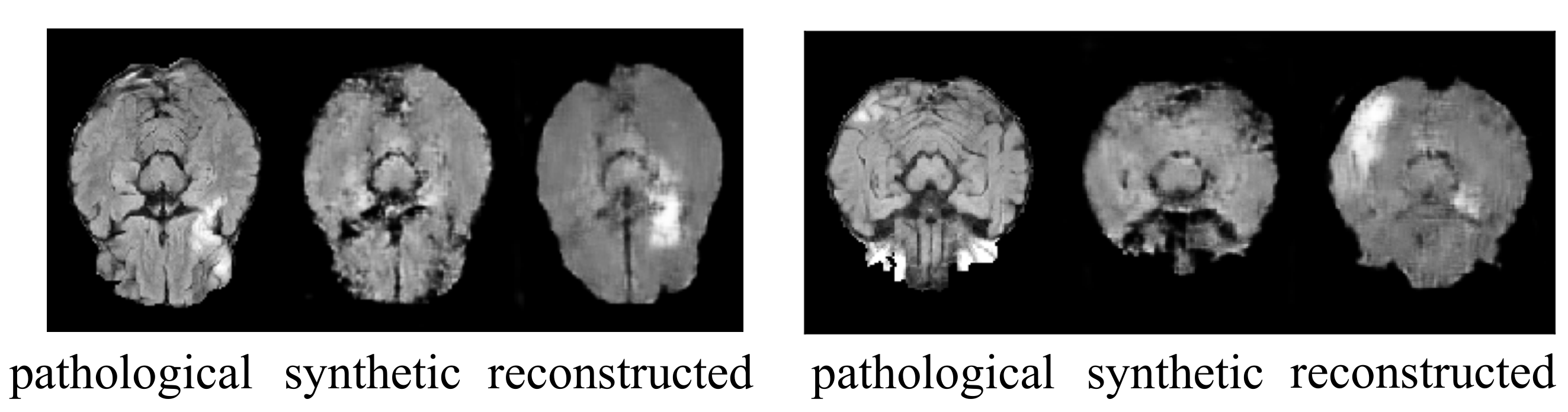}%
\caption{CycleGAN failure cases caused by the one-to-many problem. Each subfigure from left to right shows the input, the pseudo healthy and the input reconstruction. The lesion location in the reconstruction differs from the original one, since an accurate pseudo healthy image has no information to guide the reconstruction process. Images taken from ISLES.}
\label{fig: failure example}
\end{figure}
\subsection{The one-to-many problem: motivation for factorization} 
CycleGAN has to somehow invent (or hide) information when one domain contains less information than the other. In our case domain $\mathcal{P}$ does contain disease information that should not be present in $\mathcal{H}$, which leads to failure cases as shown in Figure \ref{fig: failure example}. When CycleGAN cannot invent information, \citet{chu2017cyclegan} in fact showed that it hides information within an image to be able to solve the one-to-many mapping.  Recently, several papers \cite{chartsias2018b, almahairi2018augmented, huang2018munit, lee2018diverse, esser2018variational} have shown that one needs to provide auxiliary information in the form of a style or modality specific code (actually a vector) to guide the translation and allow many-to-many mappings. Our paper does follow this practice, but instead of providing a vector we consider the auxiliary information of where the disease could be, such that the decoder does \textit{not} have to invent where things should go, and conversely the encoder does \textit{not} have to hide information.  We thus achieve that pseudo healthy images are of high quality, correspond to the identity of the same input, and also produce realistic disease maps.

\subsection{Proposed approach}
\begin{figure}[t]
\centering
\includegraphics[scale=0.4]{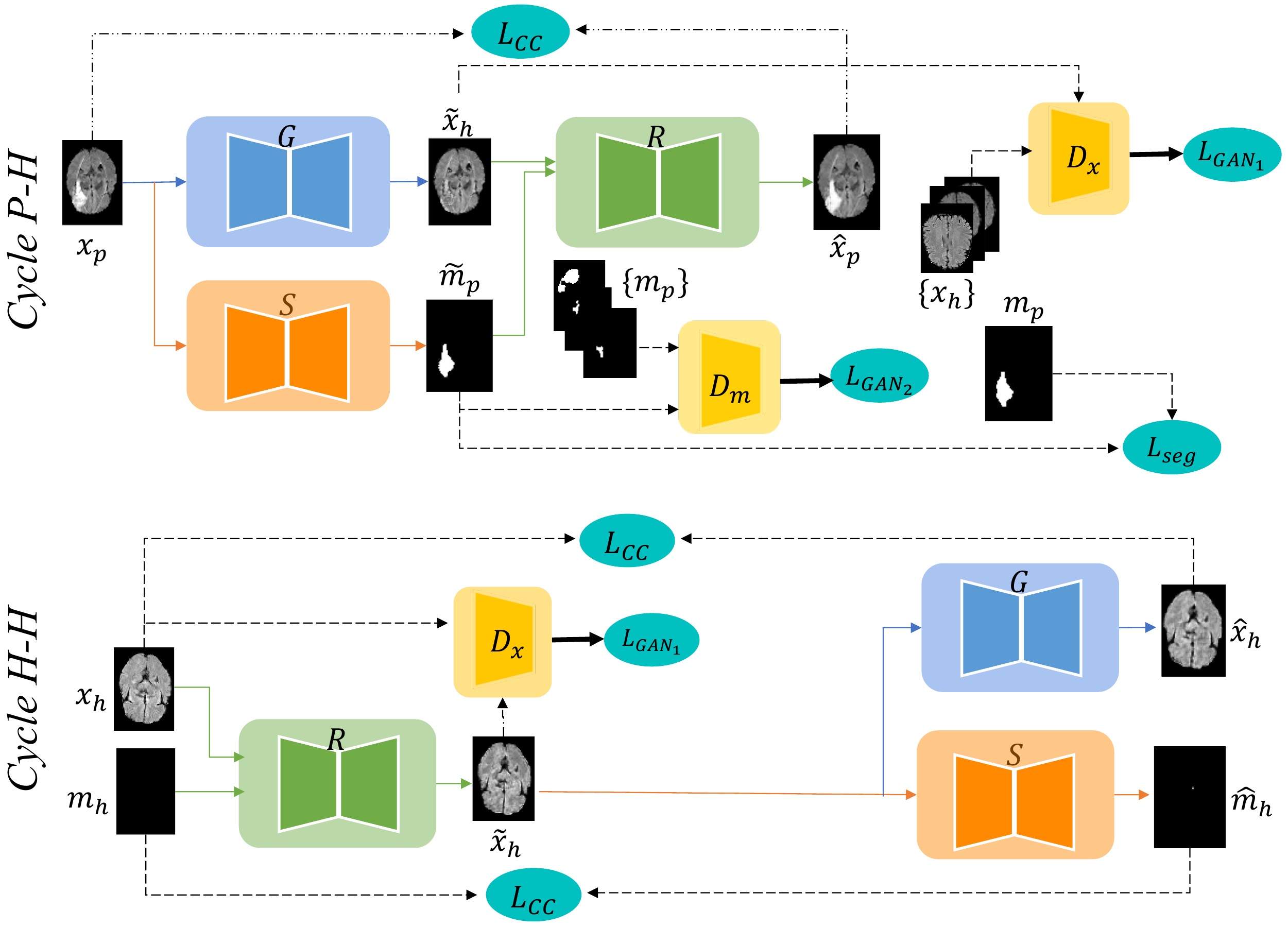}
\caption{Training flowchart. \textit{Cycle P-H} is the translation path from `pathological' to `healthy' and then back to `pathological'; \textit{Cycle H-H} is the path from a healthy image and a black mask to a fake healthy image, then back to the reconstructed image and mask.}
\label{fig: proposed_structure}
\end{figure}

 A schematic of our proposed method is illustrated in Figure \ref{fig: proposed_structure}. Recall that our task is to transform an input pathological image \textit{x\textsubscript{p}} to a disease-free image $\textit{$\mathit{\tilde{x}}$\textsubscript{h}}$ whilst maintaining the identity of \textit{x\textsubscript{p}}.
Towards this goal, our method uses the cycle-consistency losses and treats `pathological' and `healthy' as two image domains.
To solve the one-to-many mapping problem, we estimate a disease map from a pathological image using a segmentation network, and then use the map to provide information about disease location.
Specifically, there are three main components: `\textit{G}' the `generator'; `\textit{S}' the `segmentor'; and `\textit{R}' the `reconstructor' trained using two cycles: \textit{Cycle P-H} and \textit{Cycle H-H}. 

\textit{Cycle P-H}, we perform pseudo healthy synthesis, where `\textit{G}' takes a pathological image \textit{x\textsubscript{p}} as input and outputs a `healthy' looking image $\textit{$\mathit{\tilde{x}}$\textsubscript{h}}$: $\mathit{\tilde{x}\textsubscript{h} = G(x\textsubscript{p})}$. `\textit{S}' takes \textit{x\textsubscript{p}} as input and outputs a pathology map $\textit{$\mathit{\tilde{m}}$\textsubscript{p}}$: $\mathit{\tilde{m}\textsubscript{p} = S(x\textsubscript{p})}$. `\textit{R}' takes the synthesized `healthy' image $\textit{$\mathit{\tilde{x}}$\textsubscript{h}}$ and the segmented mask $\mathit{\tilde{m}_p}$ as input and outputs a `pathological' image $\textit{$\mathit{\hat{x}}$\textsubscript{p}}$: $\mathit{\hat{x}\textsubscript{p} = R(\tilde{x}\textsubscript{h}, \tilde{m}\textsubscript{p}) = R( G(x\textsubscript{p}), S(x\textsubscript{p}))} $.

\textit{Cycle H-H} utilizes healthy images and stabilizes the training. It starts with a healthy image  \textit{x\textsubscript{h}} and a null `healthy' mask \textit{m\textsubscript{h}}. First, \textit{`R'} generates a fake `healthy' image $\textit{$\mathit{\tilde{x}}$\textsubscript{h}}$: $\mathit{\tilde{x}\textsubscript{h} = R(x\textsubscript{h}, m\textsubscript{h})}$, which is then segmented into a healthy mask $\mathit{\hat{m}}\textsubscript{h}$: $\mathit{\hat{m}\textsubscript{h} = S(\tilde{x}\textsubscript{h}) = S( R(x\textsubscript{h}, m\textsubscript{h}))}$ and transformed to a reconstructed healthy image $\textit{$\mathit{\hat{x}}$\textsubscript{h}}$: $\mathit{\hat{x}\textsubscript{h} = G(\tilde{x}\textsubscript{h}) = G( R(x\textsubscript{h}, m\textsubscript{h}))}$.


There are several reasons why we design \textit{Cycle H-H} in such a way. First, because a pathology mask for a real healthy image cannot be defined.
Second, we want to prevent the reconstructor `\textit{R}' from inventing pathology when the input disease map is black. Third, we want to guide the generator \textit{`G'} and segmentor \textit{`S'} to preserve identity when the input (to both) is a `healthy' image, 
such that the synthesized `healthy' image is as similar to the input `healthy' image as possible. Similarly, when the input to the segmentor \textit{`S'} is a `healthy' image, it should output a `healthy' (no disease) map, i.e. a black mask. 

\subsection{Losses }
The training losses are $\mathcal{L\textsubscript{CC}}$, $\mathcal{L\textsubscript{GAN\textsubscript{1}}}$ and $\mathcal{L\textsubscript{Seg}}$ and $\mathcal{L\textsubscript{GAN\textsubscript{2}}}$.

$\mathbf{\mathcal{L\textsubscript{CC}}}$ is the cycle-consistency loss:
\setlength{\belowdisplayskip}{4pt} \setlength{\belowdisplayshortskip}{4pt}
\setlength{\abovedisplayskip}{4pt} \setlength{\abovedisplayshortskip}{4pt}
\begin{equation*}
\label{eqs: Lcc}
\begin{split}
    \mathcal{L\textsubscript{CC}} = \E_{x_p\sim\mathcal{P}}[\norm{ R(G(x\textsubscript{p}), S(x\textsubscript{p})) - x\textsubscript{p}}_1]  \\ + \E_{x_h\sim\mathcal{H},  m_h\sim\mathcal{H}_m}[\norm{ G(R(x\textsubscript{h}, m\textsubscript{h})) - x\textsubscript{h}}_1] + \E_{x_h\sim\mathcal{H},  m_h\sim\mathcal{H}_m}[\norm{ S(R(x\textsubscript{h}, m\textsubscript{h})) - m\textsubscript{h}}_1],
\end{split}
\end{equation*}
where the first term is defined in \textit{Cycle P-H} and the last two terms are defined in \textit{Cycle H-H}. Note that the third term uses \textit{Mean Average Error} instead of \textit{Dice}, because if the target mask is black, then given any result mask, \textit{Dice loss} will always produce 1.

$\mathbf{\mathcal{L\textsubscript{GAN\textsubscript{1}}}}$ is the least squares discriminator loss over synthetic images \cite{mao2017least}:
\begin{equation*}
\label{eqs: Lgan}
\begin{split}
    \mathcal{L\textsubscript{GAN\textsubscript{1}}} =\stackunder{max}{\textit{D}\textsubscript{x}}\,\stackunder{min}{\textit{G}}\,\frac{1}{2} \E_{x_p\sim\mathcal{P}}[\norm{ D_x(G(x\textsubscript{p})) - 1 }_2]  + \stackunder{max}{\textit{D}\textsubscript{x}}\,\frac{1}{2} \E_{x_h\sim\mathcal{H}}[\norm{ D_x(x_h)}_2] \\
    +\,\stackunder{max}{\textit{D}\textsubscript{x}}\,\stackunder{min}{\textit{R}}\,\frac{1}{2} \E_{x_h\sim\mathcal{H}, m_h\sim\mathcal{H}_m}[\norm{ D_x(R(x\textsubscript{p}, m\textsubscript{h})) - 1 }_2]
    + \stackunder{max}{\textit{D}\textsubscript{x}}\,\frac{1}{2}\E_{x_h\sim\mathcal{H}}[\norm{ D_x(x_h)}_2], 
\end{split}
\end{equation*}
where the first two terms correspond to \textit{Cycle P-H} and the last two for \textit{Cycle H-H}.

To train \textit{`S'}, we use two different training settings whether we have \textit{paired} or \textit{unpaired} data, and use a supervised or a GAN loss, respectively. 

In the \textit{paired} setting, we use manually annotated pathology masks corresponding to pathological images in $\mathcal{L\textsubscript{Seg}} = \E_{x_p\sim\mathcal{P}, m_p\sim\mathcal{P}_m}[Dice(S(x_p)-m_p)]$, with a differentiable \textit{Dice} \citep{Milletari2016VNetFC} loss. 

%

In the  \textit{unpaired} setting, since pathological images lack paired annotations, we replace $\mathcal{L\textsubscript{Seg}}$ with a discriminator \textit{D\textsubscript{m}} which classifies real pathology masks from inferred masks:
\begin{equation*}
\label{eqs: Lgan}
\begin{split}
\mathcal{L\textsubscript{GAN\textsubscript{2}}}= \stackunder{max}{\textit{D}\textsubscript{m}}\,\stackunder{min}{\textit{S}}\,\frac{1}{2} \E_{x_p\sim\mathcal{P}}[\norm{ D_m(S(x\textsubscript{p})) - 1 }_2] 
+ \stackunder{max}{\textit{D}\textsubscript{m}}\,\frac{1}{2}\E_{m_p\sim\mathcal{P}_m}[\norm{ D_m(m_p)}_2],
\end{split}
\end{equation*}
where a pathological image \textit{x\textsubscript{p}} and a mask \textit{m\textsubscript{p}} come from different volumes.

\section{Experiments}
\subsection{Experimental settings} \label{sec:experiment setting}
\textbf{Dataset and preprocessing: }
We demonstrate our method on two datasets. We use the FLAIR data of the \textit{Ischemic Lesion Segmentation} (ISLES) 2015 dataset \cite{MAIER2017250}, which contains images of 28 volumes that are skull stripped and re-sampled to an isotropic spacing of $1mm^3$ (SISS) resp. We also use FLAIR data from MRI scans of glioblastoma (GBM/HGG), made available in the \textit{Brain Tumour Segmentation} (BraTS) 2018 \cite{brats} challenge. The BraTS data contain images of 79 volumes that are skull-striped, and interpolated to $1mm^3$ resolution. Both datasets are released with segmentation masks of the pathological regions.
For each dataset, we normalize each volume by clipping the intensities to [0, $V_{99.5}$], where $V_{99.5}$ is the 99.5\% largest pixel value of the corresponding volume, then we normalize the resulting intensities to [0, 1]. We choose the middle 60 slices from each volume and label a slice as `healthy' if its corresponding pathology mask is black, and as `pathological' otherwise. We divide the datasets into a training and a testing set of 22 and 6 volumes for ISES, and 50 and 29 volumes for BraTS respectively.

\noindent \textbf{Training and implementation details:} The method is implemented in Python using Keras \cite{chollet2015keras}. The loss function for the paired data option is defined as $L_{paired}=\lambda_1 \mathcal{L\textsubscript{CC}}+ \lambda_2\mathcal{L\textsubscript{GAN\textsubscript{1}}}+\lambda_3\mathcal{L\textsubscript{Seg}}$, where $\lambda_1=10$, $\lambda_2=1$, and $\lambda_3=10$ (same values as \citet{chartsias2018b}). The loss function for the unpaired data option is defined as $L_{unpaired}=\lambda_1 \mathcal{L\textsubscript{CC}}+ \lambda_2\mathcal{L\textsubscript{GAN\textsubscript{1}}}+\lambda_3\mathcal{L\textsubscript{GAN\textsubscript{2}}}$, where $\lambda_1=10$, $\lambda_2=2$, and $\lambda_3=10$ ($\lambda_2$ has been increased to focus the attention on synthesis). Architecture details are in the Appendix.

\noindent \textbf{Baselines:} We consider two pseudo healthy synthesis baselines for comparison: a \textit{conditional GAN} \cite{mirza2014conditional} (that is deterministic and is conditioned on an image) that consists of a pseudo healthy generator, trained with unpaired data and an adversarial loss against a discriminator that classifies real and fake healthy images; and a
\textit{CycleGAN} which considers two domains for healthy and unhealthy and is trained as in \citet{zhu2017unpaired} to learn a domain translation using unpaired data.

\subsection{Evaluation metrics}
We propose, and use, numerical evaluation metrics to quantitatively evaluate the synthesized pseudo healthy images in terms of \textit{`healthiness'} and \textit{`identity'} i.e. how healthy do they look and how close to the input they are (as a proxy to identity).

\textit{`Healthiness'} is not easy to directly measure since we do not have ground-truth pseudo healthy images. However, given a pathology segmentor applied on a pseudo healthy synthetic image, we can measure the size of the segmented pathology as a proxy. To this end, we first train a segmentor to predict disease from pathological images, and then use the pre-trained segmentor to predict disease masks of synthetic pseudo healthy images and check how large the predicted disease areas are.
Formally, 
`healthiness' can be defined as:
%
%
\begin{equation*}
\textit{h}=1-\frac{\E_{\hat{x}_h\sim\mathcal{H}}[N(f\textsubscript{pre}(\hat{x}\textsubscript{h}))]}{\E_{m_p\sim\mathcal{P}_m}[N(m\textsubscript{p})]} =1- \frac{\E_{x_p\sim\mathcal{P}}[N(f\textsubscript{pre}(G(x\textsubscript{p})))]}{\E_{m_p\sim\mathcal{P}_m}[N(m\textsubscript{p})]},
\end{equation*}
where $f\textsubscript{pre}$ is the pre-trained segmentor whose output is a pathology mask, and $N(m)$ is the number of pixels which are labeled as pathology in the mask $m$. We normalize by the average size of all ground-truth pathological masks. Then we subtract the term from 1, such that $h$ increases when the images have smaller pathology.

\textit{`Identity'} is measured using a masked \textit{Multi-Scale Structural Similarity Index} (MS-SSIM) with window width 11, defined as $\text{MS-SSIM}[(1-m\textsubscript{p})\odot\hat{x}\textsubscript{h}, (1-m\textsubscript{p})\odot x\textsubscript{p}]$. This metric is based on the assumption that a pathological image and its corresponding pseudo healthy image should look the same in regions not affected by pathology. 

\subsection{Experiments on ISLES and BraTS datasets}
\begin{figure}[t]
 \centering
  \includegraphics[scale=0.49]{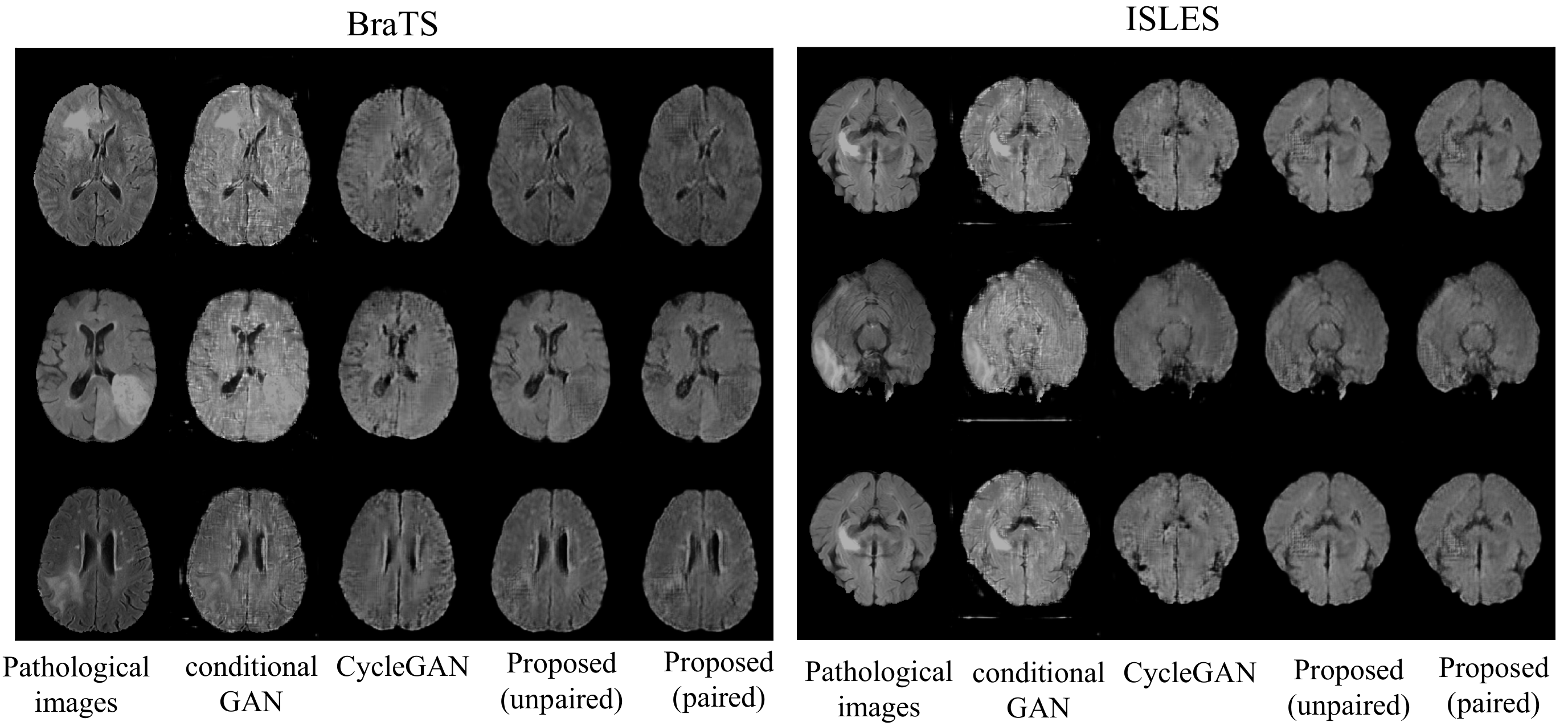}
  \caption{Experimental results for BraTS and ISLES data are shown in the \textit{left} and \textit{right} part respectively. Each part shows three samples (in three rows). The columns from left to right show the ground-truth pathological images, and pseudo healthy images generated by \textit{conditional GAN}, \textit{CycleGAN}, and the two proposed methods, respectively.  } 
  \label{fig: comparisons}
\end{figure}

We train our proposed method in both \textit{paired} and \textit{unpaired} settings on ISLES and BraTS datasets, and compare with the  baselines of Section \ref{sec:experiment setting}. Some results can be seen in Figure \ref{fig: comparisons}, where we observe that all synthetic images visually appear to be healthy. However, the pseudo healthy images generated by \textit{conditional GAN} are blurry and to some degree different from the original samples, i.e. the lateral ventricles (cavities in the middle) change: a manifestation of loss of \textit{`identity'}. Similarly, we observe changes of lateral ventricles in the synthetic images generated by \textit{CycleGAN}. These changes are probably due to the fact that \textit{CycleGAN} needs to hide information to reconstruct the input images. We also observe that our methods preserve more details of the original samples. Together, these observations imply that our proposed methods maintain better \textit{`identity'} than the baselines.

%


 We also use the proposed evaluation metrics to measure the quality of synthetic images generated by our method and baselines, respectively. The numerical results are shown in Table \ref{tab: numeric results}. We can see that our proposed method (paired) when trained using pathological image and mask pairs achieves the best results, followed by our proposed method (unpaired). Both \textit{paired} and \textit{unpaired} versions outperform conditional GAN and CycleGAN in both the BraTS and ISLES datasets. The improvements of our method are due to the factorization of pathology, which ensures maintaining information of the pathology during the pseudo healthy synthesis such that the synthetic images do not need to hide information.

\begin{table}[b]
\begin{center}
\caption{Evaluation results on BraTS and ISLES of our proposed method trained with and without \textit{pairs}, as well as of the baselines used for comparison. The best mean values for each defined metric (identity, healthiness) are shown in bold. Statistical significant results (5\% level), of our methods compared to the best baseline are marked with a star (*).
}
\begin{tabular}{r|cc|cc}
\toprule
 & \multicolumn{2}{c|}{\textbf{BraTS}}  &\multicolumn{2}{c}{\textbf{ISLES}} \\    

\textbf{Methods} & \textbf{`Identity'} & \textbf{`Healthiness'} & \textbf{`Identity'} & \textbf{`Healthiness'}\\
\midrule

 conditional GAN  & $0.74\mypm0.05$ & $0.82\mypm0.03$ & $0.67\mypm0.02$ & $0.86\mypm0.13$\\
 
 CycleGAN & $0.80\mypm0.03$  & $0.83\mypm0.04$ &$0.78\mypm0.02$ & $0.85\mypm0.11$\\
\hline 
 proposed (unpaired)  & $0.83\mypm0.03$ & $0.98\mypm0.07^*$&
 $0.82\mypm0.03$ & $0.94\mypm0.11^*$\\
 
 proposed (paired) &$\mathbf{0.88\mypm0.03}^*$ & $\mathbf{0.99\mypm}0.02^*$
&$\mathbf{0.93\mypm0.02}^*$ & $\mathbf{0.98\mypm0.04}^*$\\
 \bottomrule
\end{tabular}

\label{tab: numeric results}
\end{center}
\end{table}

\section{Conclusion}

In this paper, we propose an adversarial network for pseudo healthy synthesis with factorization of pathology. Our proposed method is composed of a pseudo healthy synthesizer to generate pseudo healthy images, a segmentor to predict a pathology map, i.e. as a way of factorizing pathology, and a reconstructor to reconstruct the input pathological image conditioned on the map. Our method can be trained in (a) \textit{paired} mode when we have paired pathological images and masks; or (b) \textit{unpaired} mode for when we do not have image and mask pairs. We also propose two numerical evaluation metrics to explicitly measure the quality of the synthesized images. We demonstrate on ISLES and BraTS datasets that our method outperforms the baselines both quantitatively and qualitatively.

Metrics that enforce or even measure identity is a topic of considerable interest in computer vision \cite{antipov2017face}. Our approach here is simple (essentially measures the fidelity of the reconstructed signal) but it does assume that changes due to disease are only local. This assumption is also adopted by several methods \cite{andermatt2018pathology, sun2018adversarial, baumgartner2017visual}. When disease globally affects an image, new approaches must be devised which is seen as future work.

\midlacknowledgments{We thank Nvidia for donating a Titan-X GPU. We also thank the University of Edinburgh and EPSRC (EP/P022928/1) for the scholarship and funding.}

\bibliography{reference.bib}

\begin{thebibliography}{31}
\providecommand{\natexlab}[1]{#1}
\providecommand{\url}[1]{\texttt{#1}}
\expandafter\ifx\csname urlstyle\endcsname\relax
  \providecommand{\doi}[1]{doi: #1}\else
  \providecommand{\doi}{doi: \begingroup \urlstyle{rm}\Url}\fi

\bibitem[Almahairi et~al.(2018)Almahairi, Rajeswar, Sordoni, Bachman, and
  Courville]{almahairi2018augmented}
Amjad Almahairi, Sai Rajeswar, Alessandro Sordoni, Philip Bachman, and Aaron~C.
  Courville.
\newblock Augmented {CycleGAN}: Learning many-to-many mappings from unpaired
  data.
\newblock In \emph{International Conference on Machine Learning}, 2018.

\bibitem[Andermatt et~al.(2018)Andermatt, Horv{\'a}th, Pezold, and
  Cattin]{andermatt2018pathology}
Simon Andermatt, Antal Horv{\'a}th, Simon Pezold, and Philippe Cattin.
\newblock {Pathology Segmentation using Distributional Differences to Images of
  Healthy Origin}.
\newblock \emph{Brain-Lesion workshop (BrainLes). MICCAI}, 2018.

\bibitem[Antipov et~al.(2017)Antipov, Baccouche, and Dugelay]{antipov2017face}
Grigory Antipov, Moez Baccouche, and Jean-Luc Dugelay.
\newblock {Face aging with conditional generative adversarial networks}.
\newblock In \emph{Image Processing (ICIP), 2017 IEEE International Conference
  on}, pages 2089--2093. IEEE, 2017.

\bibitem[Arjovsky et~al.(2017)Arjovsky, Chintala, and
  Bottou]{arjovsky2017wasserstein}
Martin Arjovsky, Soumith Chintala, and L{\'e}on Bottou.
\newblock {Wasserstein generative adversarial networks}.
\newblock In \emph{International Conference on Machine Learning}, pages
  214--223, 2017.

\bibitem[Baumgartner et~al.(2017)Baumgartner, Koch, Tezcan, Ang, and
  Konukoglu]{baumgartner2017visual}
Christian~F Baumgartner, Lisa~M Koch, Kerem~Can Tezcan, Jia~Xi Ang, and Ender
  Konukoglu.
\newblock {Visual feature attribution using Wasserstein GANs}.
\newblock In \emph{Proc IEEE Comput Soc Conf Comput Vis Pattern Recognit},
  2017.

\bibitem[Bowles et~al.(2017)Bowles, Qin, Guerrero, Gunn, Hammers, Dickie,
  Hern{\'a}ndez, Wardlaw, and Rueckert]{bowles2017brain}
Christopher Bowles, Chen Qin, Ricardo Guerrero, Roger Gunn, Alexander Hammers,
  David~Alexander Dickie, Maria~Vald{\'e}s Hern{\'a}ndez, Joanna Wardlaw, and
  Daniel Rueckert.
\newblock {Brain lesion segmentation through image synthesis and outlier
  detection}.
\newblock \emph{NeuroImage: Clinical}, 16:\penalty0 643--658, 2017.

\bibitem[Chartsias et~al.(2017)Chartsias, Joyce, Dharmakumar, and
  Tsaftaris]{chartsias2017adversarial}
Agisilaos Chartsias, Thomas Joyce, Rohan Dharmakumar, and Sotirios~A Tsaftaris.
\newblock {Adversarial image synthesis for unpaired multi-modal cardiac data}.
\newblock In \emph{International Workshop on Simulation and Synthesis in
  Medical Imaging}, pages 3--13. Springer, 2017.

\bibitem[Chartsias et~al.(2018)Chartsias, Joyce, Papanastasiou, Semple,
  Williams, Newby, Dharmakumar, and Tsaftaris]{chartsias2018b}
Agisilaos Chartsias, Thomas Joyce, Giorgos Papanastasiou, Scott Semple,
  Michelle Williams, David Newby, Rohan Dharmakumar, and Sotirios~A. Tsaftaris.
\newblock Factorised spatial representation learning: Application in
  semi-supervised myocardial segmentation.
\newblock In Alejandro~F. Frangi, Julia~A. Schnabel, Christos Davatzikos,
  Carlos Alberola-L{\'o}pez, and Gabor Fichtinger, editors, \emph{Medical Image
  Computing and Computer Assisted Intervention -- MICCAI 2018}, pages 490--498,
  Cham, 2018. Springer International Publishing.
\newblock ISBN 978-3-030-00934-2.

\bibitem[Chen and Konukoglu(2018)]{chen2018unsupervised}
Xiaoran Chen and Ender Konukoglu.
\newblock {Unsupervised Detection of Lesions in Brain MRI using constrained
  adversarial auto-encoders}.
\newblock \emph{Internatinal Conference on Medical Imaging with Deep Learning},
  2018.

\bibitem[Chollet et~al.(2015)]{chollet2015keras}
Fran\c{c}ois Chollet et~al.
\newblock Keras.
\newblock \url{https://keras.io}, 2015.

\bibitem[Chu et~al.(2017)Chu, Zhmoginov, and Sandler]{chu2017cyclegan}
Casey Chu, Andrey Zhmoginov, and Mark Sandler.
\newblock {CycleGAN: a Master of Steganography}.
\newblock \emph{NIPS 2017, Workshop on Machine Deception}, 2017.

\bibitem[Esser et~al.(2018)Esser, Sutter, and Ommer]{esser2018variational}
Patrick Esser, Ekaterina Sutter, and Bj{\"o}rn Ommer.
\newblock {A Variational U-Net for Conditional Appearance and Shape
  Generation}.
\newblock In \emph{Proceedings of the IEEE Conference on Computer Vision and
  Pattern Recognition}, pages 8857--8866, 2018.

\bibitem[Frangi et~al.(2018)Frangi, Tsaftaris, and Prince]{special2018Frangi}
A.~F. Frangi, S.~A. Tsaftaris, and J.~L. Prince.
\newblock {Simulation and Synthesis in Medical Imaging}.
\newblock \emph{IEEE Transactions on Medical Imaging}, 37\penalty0
  (3):\penalty0 673--679, March 2018.
\newblock ISSN 0278-0062.
\newblock \doi{10.1109/TMI.2018.2800298}.

\bibitem[Huang et~al.(2018)Huang, Liu, Belongie, and Kautz]{huang2018munit}
Xun Huang, Ming-Yu Liu, Serge Belongie, and Jan Kautz.
\newblock Multimodal unsupervised image-to-image translation.
\newblock In \emph{European Conference on Computer Vision}, volume 11207, pages
  179--196. Springer International Publishing, 2018.

\bibitem[Huo et~al.(2018)Huo, Xu, Moon, Bao, Assad, Moyo, Savona, Abramson, and
  Landman]{huo2018}
Y.~Huo, Z.~Xu, H.~Moon, S.~Bao, A.~Assad, T.~K. Moyo, M.~R. Savona, R.~G.
  Abramson, and B.~A. Landman.
\newblock {SynSeg-Net: Synthetic Segmentation Without Target Modality Ground
  Truth}.
\newblock \emph{IEEE Transactions on Medical Imaging}, pages 1--1, 2018.
\newblock ISSN 0278-0062.
\newblock \doi{10.1109/TMI.2018.2876633}.

\bibitem[Lee et~al.(2018)Lee, Tseng, Huang, Singh, and Yang]{lee2018diverse}
Hsin-Ying Lee, Hung-Yu Tseng, Jia-Bin Huang, Maneesh~Kumar Singh, and
  Ming-Hsuan Yang.
\newblock {Diverse Image-to-Image Translation via Disentangled
  Representations}.
\newblock In \emph{European Conference on Computer Vision}, volume 11205, pages
  36--52. Springer International Publishing, 2018.

\bibitem[Maier et~al.(2017)Maier, Menze, von~der Gablentz, Häni, Heinrich,
  Liebrand, Winzeck, Basit, Bentley, Chen, Christiaens, Dutil, Egger, Feng,
  Glocker, Götz, Haeck, Halme, Havaei, Iftekharuddin, Jodoin, Kamnitsas,
  Kellner, Korvenoja, Larochelle, Ledig, Lee, Maes, Mahmood, Maier-Hein,
  McKinley, Muschelli, Pal, Pei, Rangarajan, Reza, Robben, Rueckert, Salli,
  Suetens, Wang, Wilms, Kirschke, Krämer, Münte, Schramm, Wiest, Handels, and
  Reyes]{MAIER2017250}
Oskar Maier, Bjoern~H. Menze, Janina von~der Gablentz, Levin Häni, Mattias~P.
  Heinrich, Matthias Liebrand, Stefan Winzeck, Abdul Basit, Paul Bentley, Liang
  Chen, Daan Christiaens, Francis Dutil, Karl Egger, Chaolu Feng, Ben Glocker,
  Michael Götz, Tom Haeck, Hanna-Leena Halme, Mohammad Havaei, Khan~M.
  Iftekharuddin, Pierre-Marc Jodoin, Konstantinos Kamnitsas, Elias Kellner,
  Antti Korvenoja, Hugo Larochelle, Christian Ledig, Jia-Hong Lee, Frederik
  Maes, Qaiser Mahmood, Klaus~H. Maier-Hein, Richard McKinley, John Muschelli,
  Chris Pal, Linmin Pei, Janaki~Raman Rangarajan, Syed~M.S. Reza, David Robben,
  Daniel Rueckert, Eero Salli, Paul Suetens, Ching-Wei Wang, Matthias Wilms,
  Jan~S. Kirschke, Ulrike~M. Krämer, Thomas~F. Münte, Peter Schramm, Roland
  Wiest, Heinz Handels, and Mauricio Reyes.
\newblock {"ISLES 2015 - A public evaluation benchmark for ischemic stroke
  lesion segmentation from multispectral MRI"}.
\newblock \emph{Medical Image Analysis}, 35:\penalty0 250 -- 269, 2017.
\newblock ISSN 1361-8415.
\newblock \doi{https://doi.org/10.1016/j.media.2016.07.009}.

\bibitem[Mao et~al.(2017)Mao, Li, Xie, Lau, Wang, and Smolley]{mao2017least}
Xudong Mao, Qing Li, Haoran Xie, Raymond~YK Lau, Zhen Wang, and Stephen~Paul
  Smolley.
\newblock {Least squares generative adversarial networks}.
\newblock In \emph{Computer Vision (ICCV), 2017 IEEE International Conference
  on}, pages 2813--2821. IEEE, 2017.

\bibitem[Menze et~al.(2015)Menze, Jakab, Bauer, Kalpathy-Cramer, Farahani,
  Kirby, Burren, Porz, Slotboom, Wiest, Lanczi, Gerstner, Weber, Arbel, Avants,
  Ayache, Buendia, Collins, Cordier, Corso, Criminisi, Das, Delingette,
  Demiralp, Durst, Dojat, Doyle, Festa, Forbes, Geremia, Glocker, Golland, Guo,
  Hamamci, Iftekharuddin, Jena, John, Konukoglu, Lashkari, Mariz, Meier,
  Pereira, Precup, Price, Raviv, Reza, Ryan, Sarikaya, Schwartz, Shin, Shotton,
  Silva, Sousa, Subbanna, Szekely, Taylor, Thomas, Tustison, Unal, Vasseur,
  Wintermark, Ye, Zhao, Zhao, Zikic, Prastawa, Reyes, and Leemput]{brats}
B.~H. Menze, A.~Jakab, S.~Bauer, J.~Kalpathy-Cramer, K.~Farahani, J.~Kirby,
  Y.~Burren, N.~Porz, J.~Slotboom, R.~Wiest, L.~Lanczi, E.~Gerstner, M.~Weber,
  T.~Arbel, B.~B. Avants, N.~Ayache, P.~Buendia, D.~L. Collins, N.~Cordier,
  J.~J. Corso, A.~Criminisi, T.~Das, H.~Delingette, Ç. Demiralp, C.~R. Durst,
  M.~Dojat, S.~Doyle, J.~Festa, F.~Forbes, E.~Geremia, B.~Glocker, P.~Golland,
  X.~Guo, A.~Hamamci, K.~M. Iftekharuddin, R.~Jena, N.~M. John, E.~Konukoglu,
  D.~Lashkari, J.~A. Mariz, R.~Meier, S.~Pereira, D.~Precup, S.~J. Price, T.~R.
  Raviv, S.~M.~S. Reza, M.~Ryan, D.~Sarikaya, L.~Schwartz, H.~Shin, J.~Shotton,
  C.~A. Silva, N.~Sousa, N.~K. Subbanna, G.~Szekely, T.~J. Taylor, O.~M.
  Thomas, N.~J. Tustison, G.~Unal, F.~Vasseur, M.~Wintermark, D.~H. Ye,
  L.~Zhao, B.~Zhao, D.~Zikic, M.~Prastawa, M.~Reyes, and K.~Van Leemput.
\newblock {The Multimodal Brain Tumor Image Segmentation Benchmark (BRATS)}.
\newblock \emph{IEEE Transactions on Medical Imaging}, 34\penalty0
  (10):\penalty0 1993--2024, Oct 2015.
\newblock ISSN 0278-0062.
\newblock \doi{10.1109/TMI.2014.2377694}.

\bibitem[Milletari et~al.(2016)Milletari, Navab, and
  Ahmadi]{Milletari2016VNetFC}
Fausto Milletari, Nassir Navab, and Seyed-Ahmad Ahmadi.
\newblock {{V-Net}: Fully Convolutional Neural Networks for Volumetric Medical
  Image Segmentation}.
\newblock \emph{2016 Fourth International Conference on 3D Vision}, pages
  565--571, 2016.
\newblock \doi{10.1109/3DV.2016.79}.

\bibitem[Mirza and Osindero(2014)]{mirza2014conditional}
Mehdi Mirza and Simon Osindero.
\newblock {Conditional generative adversarial nets}.
\newblock \emph{arXiv preprint arXiv:1411.1784}, 2014.

\bibitem[Ronneberger et~al.(2015)Ronneberger, Fischer, and
  Brox]{ronneberger2015u}
Olaf Ronneberger, Philipp Fischer, and Thomas Brox.
\newblock {U-net: Convolutional networks for biomedical image segmentation}.
\newblock In \emph{International Conference on Medical image computing and
  computer-assisted intervention}, pages 234--241. Springer, 2015.

\bibitem[Schlegl et~al.(2017)Schlegl, Seeb{\"o}ck, Waldstein, Schmidt-Erfurth,
  and Langs]{schlegl2017unsupervised}
Thomas Schlegl, Philipp Seeb{\"o}ck, Sebastian~M Waldstein, Ursula
  Schmidt-Erfurth, and Georg Langs.
\newblock {Unsupervised anomaly detection with generative adversarial networks
  to guide marker discovery}.
\newblock In \emph{International Conference on Information Processing in
  Medical Imaging}, pages 146--157. Springer, 2017.

\bibitem[Sun et~al.(2018)Sun, Wang, Ding, Huang, and
  Paisley]{sun2018adversarial}
Liyan Sun, Jiexiang Wang, Xinghao Ding, Yue Huang, and John Paisley.
\newblock {An Adversarial Learning Approach to Medical Image Synthesis for
  Lesion Removal}.
\newblock \emph{arXiv preprint arXiv:1810.10850}, 2018.

\bibitem[Tsunoda et~al.(2014)Tsunoda, Moribe, Orii, Kawano, and
  Maeda]{tsunoda2014pseudo}
Yuriko Tsunoda, Masayuki Moribe, Hideaki Orii, Hideaki Kawano, and Hiroshi
  Maeda.
\newblock {Pseudo-normal image synthesis from chest radiograph database for
  lung nodule detection}.
\newblock In \emph{Advanced Intelligent Systems}, pages 147--155. Springer,
  2014.

\bibitem[Wang et~al.(2018)Wang, Macnaught, Papanastasiou, MacGillivray, and
  Newby]{wang2018unsupervised}
Chengjia Wang, Gillian Macnaught, Giorgos Papanastasiou, Tom MacGillivray, and
  David Newby.
\newblock {Unsupervised learning for cross-domain medical image synthesis using
  deformation invariant cycle consistency networks}.
\newblock In \emph{International Workshop on Simulation and Synthesis in
  Medical Imaging}, pages 52--60. Springer, 2018.

\bibitem[Wolterink et~al.(2017)Wolterink, Dinkla, Savenije, Seevinck, van~den
  Berg, and I{\v{s}}gum]{wolterink2017deep}
Jelmer~M Wolterink, Anna~M Dinkla, Mark~HF Savenije, Peter~R Seevinck,
  Cornelis~AT van~den Berg, and Ivana I{\v{s}}gum.
\newblock {Deep MR to CT synthesis using unpaired data}.
\newblock In \emph{International Workshop on Simulation and Synthesis in
  Medical Imaging}, pages 14--23. Springer, 2017.

\bibitem[Yang et~al.(2016)Yang, Han, Park, Aylward, Kwitt, and
  Niethammer]{yang2016registration}
Xiao Yang, Xu~Han, Eunbyung Park, Stephen Aylward, Roland Kwitt, and Marc
  Niethammer.
\newblock {Registration of pathological images}.
\newblock In \emph{International Workshop on Simulation and Synthesis in
  Medical Imaging}, pages 97--107. Springer, 2016.

\bibitem[Ye et~al.(2013)Ye, Zikic, Glocker, Criminisi, and
  Konukoglu]{ye2013modality}
Dong~Hye Ye, Darko Zikic, Ben Glocker, Antonio Criminisi, and Ender Konukoglu.
\newblock {Modality propagation: coherent synthesis of subject-specific scans
  with data-driven regularization}.
\newblock In \emph{International Conference on Medical Image Computing and
  Computer-Assisted Intervention}, pages 606--613. Springer, 2013.

\bibitem[Zhang et~al.(2018)Zhang, Yang, and Zheng]{zhang2018translating}
Zizhao Zhang, Lin Yang, and Yefeng Zheng.
\newblock {Translating and segmenting multimodal medical volumes with cycle-and
  shapeconsistency generative adversarial network}.
\newblock In \emph{Proceedings of the IEEE Conference on Computer Vision and
  Pattern Recognition}, pages 9242--9251, 2018.

\bibitem[Zhu et~al.(2017)Zhu, Park, Isola, and Efros]{zhu2017unpaired}
Jun-Yan Zhu, Taesung Park, Phillip Isola, and Alexei~A Efros.
\newblock {Unpaired Image-to-Image Translation using Cycle-Consistent
  Adversarial Networks}.
\newblock In \emph{IEEE International Conference on Computer Vision}, 2017.

\end{thebibliography}

\appendix
\section{Architecture details}
The detailed architecture of our generator \textit{`G'} is shown in Table \ref{tab:generator}. IN stands for Instance Normalization. The detailed architecture of our reconstructor \textit{`R'} is shown in Table \ref{tab:reconstructor}. The detailed architecture of our discriminator `$\textit{D}_x$' and `$\textit{D}_m$' is shown in Table \ref{tab:discriminator}.

\begin{table}[h!]
    \centering
    \begin{tabular}{|c|c|c|c|c|c|c|}
        Layer & Input & filter size & stride & IN & activation & Output\\
        \hline
        conv2d & (208,160,1) & 7 &1 & Yes & ReLu & (208,160,32) \\
        conv2d & (208,160,32) & 3 &2 & Yes &ReLu & (104,80,64) \\
        conv2d & (104,80,64)  & 3 &2 & Yes &ReLu & (52,40,128) \\
        residual block &  (52,40,128) &3 &1 &Yes & Leaky ReLu&(52,40,128)\\
        residual block &  (52,40,128) &3 &1 &Yes & Leaky ReLu&(52,40,128)\\
        residual block &  (52,40,128) &3 &1 &Yes & Leaky ReLu&(52,40,128)\\
        residual block &  (52,40,128) &3 &1 &Yes & Leaky ReLu&(52,40,128)\\
        residual block &  (52,40,128) &3 &1 &Yes & Leaky ReLu&(52,40,128)\\
        residual block &  (52,40,128) &3 &1 &Yes & Leaky ReLu&(52,40,128)\\   
        upsampling2d & (52,40,128) & - & - & - & - & (104, 80, 128)\\
        conv2d & (104, 80, 128) & 3 & 1& Yes & ReLu & (104,80,64)\\
        upsampling2d & (104,80,64) & - & - & - & - & (208, 160, 64)\\
        conv2d & (208, 160, 64) & 3 & 1& Yes & ReLu & (208, 160, 32)\\
        
        conv2d & (208, 160, 32) & 3 & 1& No & sigmoid & (208, 160, 1)\\
        
    \end{tabular}
    \caption{Detailed architecture of generator \textit{`G'}.}
    \label{tab:generator}
\end{table}

\begin{table}[h!]
    \centering
    \begin{tabular}{|c|c|c|c|c|c|c|}
        Layer & Input & filter size & stride & IN & activation & Output\\
        \hline
        conv2d & (208,160,2) & 7 &1 & Yes & ReLu & (208,160,32) \\
        conv2d & (208,160,32) & 3 &2 & Yes &ReLu & (104,80,64) \\
        conv2d & (104,80,64)  & 3 &2 & Yes &ReLu & (52,40,128) \\
        residual block &  (52,40,128) &3 &1 &Yes & Leaky ReLu&(52,40,128)\\
        residual block &  (52,40,128)&3 &1 &Yes & Leaky ReLu&(52,40,128)\\
        residual block &  (52,40,128) &3 &1 &Yes & Leaky ReLu&(52,40,128)\\
        residual block &  (52,40,128) &3 &1 &Yes & Leaky ReLu&(52,40,128)\\
        residual block &  (52,40,128) &3 &1 &Yes & Leaky ReLu&(52,40,128)\\
        residual block &  (52,40,128) &3 &1 &Yes & Leaky ReLu&(52,40,128)\\   
        upsampling2d & (52,40,128) & - & - & - & - & (104, 80, 128)\\
        conv2d & (104, 80, 128) & 3 & 1& Yes & ReLu & (104,80,64)\\
        upsampling2d & (104,80,64) & - & - & - & - & (208, 160, 64)\\
        conv2d & (208, 160, 64) & 3 & 1& Yes & ReLu & (208, 160, 32)\\
        conv2d & (208, 160, 32) & 3 & 1& No & sigmoid & (208, 160, 2)\\
    \end{tabular}
    \caption{Detailed architecture of reconstructor \textit{`R'}.}
    \label{tab:reconstructor}
\end{table}

\begin{table}[h!]
    \centering
    \begin{tabular}{|c|c|c|c|c|c|c|}
        Layer & Input & filter size & stride & IN & activation & Output\\
        \hline
        conv2d & (208,160,2) & 4 &2 & Yes & Leaky ReLu & (104,80,32) \\
        conv2d & (104,80,32) & 4 &2 & Yes &Leaky ReLu & (52,40,128) \\
        conv2d & (52,40,128)  & 4 &2 & Yes &Leaky ReLu & (26,20,256) \\
        conv2d & (26,20,256)  & 4 &2 & Yes &Leaky ReLu & (13,10,512) \\
        conv2d &  (13,10,512)  & 4 &1 & No &sigmoid & (13,10,1)
    \end{tabular}
    \caption{Detailed architecture of discriminator `$\textit{D}_x$' and `$\textit{D}_m$'.}
    \label{tab:discriminator}
\end{table}

\newpage
The detailed architecture of our segmentor $\textit{`S'}$ is a U-Net, and follows the structure of \citet{ronneberger2015u}. We change the activation function from `ReLu' to `Leaky ReLu'. We also found that using residual connection on each layer slightly improved the results.

The pre-trained segmentor $\mathit{f_{pre}}$ which is used for evaluation uses the same structure as $\textit{`S'}$. We train the segmentor $\mathit{f_{pre}}$ on the ISLES and BraTS training datasets (see Section 4.1) respectively, and then use it to evaluate synthetic images generated from samples in ISLES and BraTS testing datasets. The Dice loss of the segmentor on ISLES and BraTS testing datasets are 0.12 and 0.16, respectively.

\end{document}